\documentclass[11pt,a4paper]{article}
\usepackage[hyperref]{acl2020}
\usepackage{times}
\usepackage{latexsym}

\usepackage{microtype}
\usepackage{booktabs}
\usepackage{makecell}
\usepackage{tabularx}
\usepackage{array}
\newcolumntype{C}{>{\centering\arraybackslash}p{4.7em}}
\usepackage{multirow}
\usepackage{colortbl}
\usepackage{amsmath}
\usepackage{amsfonts}
\usepackage{graphicx}
\usepackage{subcaption}
\usepackage{url}

\aclfinalcopy 

\title{Towards Instance-Level Parser Selection for \\ Cross-Lingual Transfer of Dependency Parsers}

\author{Robert Litschko$^{\mathbf{1}}$, ~ Ivan Vuli\'{c}$^{\mathbf{2}}$, ~ \v{Z}eljko Agi\'{c}$^{\mathbf{3}}$, ~ {Goran Glava\v{s}}$^{\mathbf{1}}$\\
$^{\mathbf{1}}$ Data and Web Science Group, University of Mannheim, Germany \\
$^{\mathbf{2}}$ Language Technology Lab, University of Cambridge, UK \\
$^{\mathbf{3}}$ Corti, Copenhagen, Denmark\\
\texttt{\{litschko,goran\}@informatik.uni-mannheim.de} \\ \texttt{iv250@cam.ac.uk} \hspace{0.5em} \texttt{za@corti.ai}}

\date{}

\begin{document}
\maketitle
\begin{abstract}
Current methods of cross-lingual parser transfer focus on predicting the best parser for a low-resource target language globally, that is, ``at treebank level''. In this work, we propose and argue for a novel cross-lingual transfer paradigm: \textit{instance-level parser selection} (ILPS), and present a proof-of-concept study focused on instance-level selection in the framework of \textit{delexicalized parser transfer}. We start from an empirical observation that different source parsers are the best choice for different Universal POS sequences in the target language. We then propose to predict the best parser at the instance level. To this end, we train a supervised regression model, based on the Transformer architecture, to predict parser accuracies for individual POS-sequences. We compare ILPS against two strong single-best parser selection baselines (SBPS): (1) a model that compares POS n-gram distributions between the source and target languages (KL) and (2) a model that selects the source based on the similarity between manually created language vectors encoding syntactic properties of languages (L2V). The results from our extensive evaluation, coupling 42 source parsers and 20 diverse low-resource test languages, show that ILPS outperforms KL and L2V on 13/20 and 14/20 test languages, respectively. Further, we show that by predicting the best parser ``at the treebank level'' (SBPS), using the aggregation of predictions from our instance-level model, we outperform the same baselines on 17/20 and 16/20 test languages.
\end{abstract}

\section{Introduction}

 A major goal and promise of cross-lingual transfer in NLP is to transfer language technology to as many languages as possible \cite{ohoran-etal-2016-survey,Ponti:2019cl}. Proper language-specific automated syntactic analyses are still unavailable for 99\% of the world's languages due to the lack of respective treebanks. Therefore, delexicalized transfer of dependency parsers has profiled as the most viable cross-lingual transfer option \cite{zeman-resnik-2008-cross,mcdonald-etal-2011-multi,sogaard2011data}. Delexicalized transfer is conceptually the least demanding option in terms of language-specific resource requirements. The only provision, in order to transfer the parser trained on a delexicalized treebank of a resource-rich language, is a POS tagger in a low-resource target language based on the Universal POS (UPOS) tagset \cite{petrov2012universal}. Delexicalized transfer is nowadays used primarily as a simple yet competitive baseline for more sophisticated transfer models. However, in realistic low-resource setups, one cannot guarantee additional resources such as parallel sentences \cite{ma2014unsupervised,rasooli2015density,rasooli2017cross}, word alignments \cite{lacroix2016frustratingly}, or even sufficiently large monolingual corpus in the target language \cite{mulcaire-etal-2019-low}. Thus, delexicalized transfer remains a widely useful and plausible option \cite{johannsen2016joint,agic2017cross}.

Delexicalized transfer comes in two main flavors. We either (1) choose the best parser from a set of available parsers, trained on treebanks of various resource-rich languages (\textit{single-best parser selection}, SBPS) or (2) use the parser trained on a mixture of treebanks of (ideally related) resource-rich languages (\textit{multi-source parser transfer}, MSP). Other transfer paradigms, like noise-based data augmentation \cite{sahin-steedman-2018-data}, assume the existence of at least a small treebank for a target language, violating the assumption of a (treebank-wise) fully low-resource target language. 
%

Both SBPS and MSP rely on some measure of structural alignment between languages in order to select either the single best source language parser (SBPS) or a set of (syntactically related) source languages (MSP). Existing solutions rely on measures like the Kullback–Leibler (KL) divergence between source- and target-language distributions of POS trigrams \cite{rosa-zabokrtsky-2015-klcpos3}, which can be unreliable for small target language corpora or instance-level estimation. More recent approaches \cite{agic2017cross,lin-etal-2019-choosing} choose suitable source languages based on manually coded typological similarities between languages available from databases such as WALS \cite{dryer2013online} or URIEL \cite{littell2017uriel}. The bottleneck of this approach is the manual effort and linguistic expertise needed to introduce a new language into the database.

\vspace{1.6mm}
\noindent \textbf{Proof-of-Concept and Contributions.} In this work, we propose a novel paradigm for cross-lingual parser transfer. The idea is to select the source-language parser for each target instance (i.e., POS-sequence), dubbed \textit{instance-level parser selection} (ILPS). This is motivated by a simple observation that different source parsers provide most accurate parses for different target POS-sequences. We empirically show that an oracle ILPS leads to major potential gains compared to an oracle single-best parser selection at the treebank level (SBPS). 

As a proof-of-concept for ILPS, we present a neural regression model that predicts the accuracy of a source-language parser for a given UPOS-``sentence'' (i.e., a sequence of universal POS tags). We measure accuracies of parsers of resource-rich languages on UPOS-sentences from treebanks of other resource-rich languages to create training examples for the regression model. At inference time, we apply the trained regression model to select the best parser \textit{for each instance} (i.e., each UPOS-sentence) of a low-resource target language.\footnote{In contrast to existing methods which impose additional requirements on the target language (e.g., a sufficiently large target language corpus or an expert linguistic specification of the language's syntactic properties), our ILPS setting conforms to a more realistic minimal-resource setup: it does not rely on any target-language resource other than a POS tagger.}

We perform a large-scale evaluation of delexicalized dependency parser transfer, encompassing 42 source languages with large(r) treebanks, and 20 target (i.e., test) languages with small(er) treebanks from the Universal Dependencies (UD) v2.3 collection \cite{nivre2018ud23}. We show that, averaged across all test treebanks, our simple ILPS model significantly outperforms strong SBPS baselines \cite{rosa-zabokrtsky-2015-klcpos3,lin-etal-2019-choosing}. We further demonstrate that we can easily aggregate instance-level predictions into an SBPS model, yielding improvements over the existing SBPS baselines for 16/20 and 17/20 test languages. Finally, we show that by ensembling the parses of few-best parsers according to the ILPS model's predictions we can significantly outperform (1) the multi-source parser trained on the treebanks of all 42 source languages and (2) even surpass the performance of an oracle single-best treebank-level parser selection (i.e., oracle SBPS). 

We believe that this \textit{proof-of-concept work} uncovers the great potential of instance-based parser selection for cross-lingual parsing transfer for truly low-resource setups. The gaps with respect to the oracle (upper-bound) ILPS performance indicate that we have only scratched the surface of this potential. We hope that our work will inspire further investigations of this promising cross-lingual transfer paradigm in other setups and for other tasks.

\section{Motivation: The Case for Instance-Based Parser Selection}
\label{sec:motivation}

The idea behind instance-level parser selection is intuitive: given a set of parsers for resource-rich source languages, it is unlikely that the same source-language parser is the best choice for all instances (i.e., UPOS-sentences) of the target language. Therefore, we first investigate the performance of an oracle model that would be able to predict the best source-language parser for each individual POS-sentence from the target-language treebank. 
To verify this, we rely on the well-known biaffine parser \cite{dozat2017deep,dozat-etal-2017-stanfords} and train it on delexicalized UD2.3 treebanks \cite{nivre2018ud23} of 42 languages.\footnote{We selected 42 languages with largest treebanks as the training languages. For languages with multiple treebanks (e.g., \textsc{en}, \textsc{cs}), we finally chose the treebank for which the parser yielded the best monolingual parsing accuracy.} We then parse the delexicalized treebanks of the 20 low-resource languages with all 42 source parsers, and measure their performance per each instance in each target treebank. We compare the performance of two \textit{oracle} parser selection strategies: (1) single-best parser selection (SBPS), in which for each target test treebank we select the parser that performs best on the entire treebank; and (2) instance-level parser selection strategy (ILPS), where for each UPOS-sentence from each test treebank, we select the parser that produces the best parse for that UPOS-sentence.

\begin{figure*}
    \begin{center}
    \includegraphics[scale=0.52]{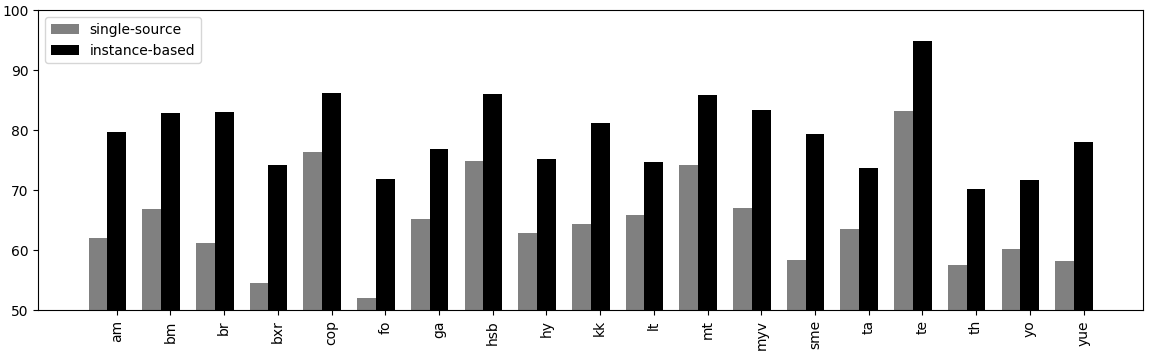}
    \caption{Comparison of UAS between \textit{oracle} single-best (i.e., treebank-level) parser selection (SBPS) and instance-level parser selection (ILPS) strategies for cross-lingual transfer of delexicalized parsers for 20 low-resource languages from UD2.3, used as test languages throughout the paper.}
    \label{fig:potential_gains}
    \end{center}
    \vspace{-2mm}
\end{figure*}


The differences in Unlabeled Attachment Scores (UAS) between the two transfer paradigms are shown in Figure~\ref{fig:potential_gains}. This clearly demonstrates a large gap in favor of ILPS: the average gain with ILPS is 14.5 UAS points, and it is prominent for all languages. It suggests that large improvements may be obtained with a model that can predict the best parser at the instance level, that is, for each UPOS-sentence separately. However, these are still oracle scores and we pose the following research question in this paper: \textit{(Q1) Is it possible to learn an instance-level prediction model to select the best parser given any UPOS-sentence, irrespective to its ``language of origin''?}\footnote{Note that in theory the \textit{oracle} gaps in favor of ILPS may be out of reach for automatic ILPS models, due to a potential parsing ambiguity introduced through delexicalization -- i.e., the same UPOS-sentence (corresponding to different lexicalized sentences) may appear in the same treebank or across different treebanks with different gold parses. However, we have verified that this phenomenon is rare: ambiguous parses are present only for 1.4\% UPOS-sentences in the concatenation of treebanks from 42 languages.} In addition, even with noisy automatic instance-level predictions, one could still, by eliminating the noise through aggregation, use them to inform treebank-level source parser selection. In other words, another research question we pose is: \textit{(Q2) Can we improve single-best global parser selection through aggregating instance-level parser predictions?} 

\section{Instance-Based Parser Selection}

\begin{figure*}[t]
\centering
\includegraphics[width=1.0\textwidth]{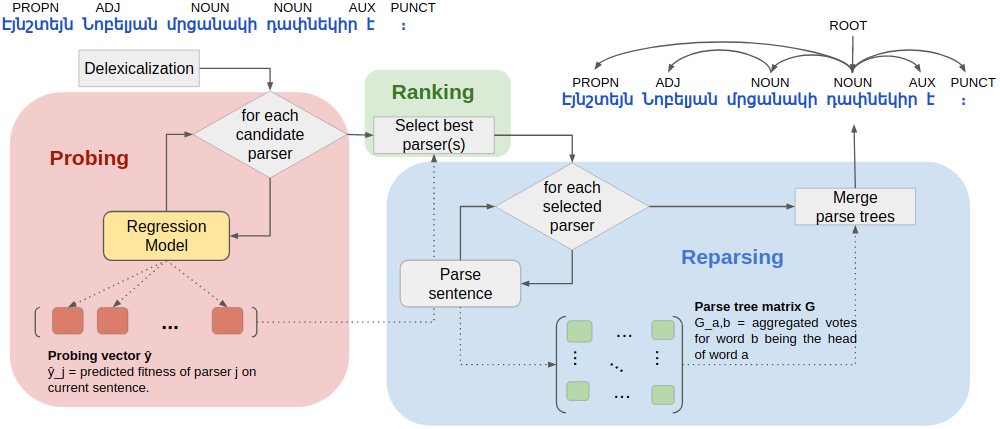}
\caption{Illustration of the ILPS framework (at inference time) with three steps using an example sentence in Armenian (\textsc{hy}): (1) Probing -- the ILPS regression model predicts the parsing accuracy on a given test UPOS-sentence for each of the 42 parsers; (2) Ranking -- rank the parsers w.r.t. parsing accuracy for the instance and selects one or few best-performing parsers; (3) Reparsing -- induce the final tree for the UPOS-sentence by merging trees produced by parsers selected in the previous step (only if more than one parser gets selected in step (2)).}
\label{fig:overview}
\end{figure*}

We now describe a novel ILPS framework based on a supervised regression model that predicts the parser accuracy for any UPOS-sentence. As such, it can be applied on UPOS-sequences of low-resource languages. As described in \S\ref{sec:motivation}, we first train a (biaffine) parser on delexicalized treebanks for each of the 42 resource-rich languages from UD2.3.\footnote{All language codes used throughout this paper are taken directly from the UD2.3 documentation \cite{nivre2018ud23}.} We then parse with each parser the 41 treebanks of the other languages. This way we obtain the labels for training the ILPS regression model.  The data preparation step is further detailed in \S\ref{sec:prep_tr_data}, while the model is described in \S\ref{ss:model}. At inference time, the ILPS model predicts the accuracy of each of the 42 parsers for each UPOS-sentence from delexicalized treebanks of the 20 test languages.\footnote{Note that this constitutes a minimal-resource transfer setup: our ILPS regression model does not rely on any information about the test languages nor their respective treebanks.} Finally, in \S\ref{sec:ranking} and \S\ref{sec:reparsing}, we outline different strategies for merging the parse trees based on the predictions of the ILPS regression model. The full ILPS framework is illustrated in Figure~\ref{fig:overview}.

\subsection{Preparing ILPS Training Data}
\label{sec:prep_tr_data}

We first delexicalize all treebanks before training the parsers. After training a parser for each of the $|L|$ training languages, we measure how each of them performs on treebanks of the other $|L|-1$ languages. Let $\text{PARSER}_{i}$ denote the parser of the $i$-th training language and let $\text{SENT}_j = \{\text{POS}\}_{n = 1}^{N}$ be an UPOS-sentence of length $N$ from the treebank of the $j$-th training language. Next, we must quantify how successful $\text{PARSER}_{i}$ is on some UPOS-sentence $\text{SENT}_j$. To this end, we use the number of correct dependency heads predicted by $\text{PARSER}_{i}$ on $\text{SENT}_j$. Using a raw number of correct heads as training labels for the ILPS regression model comes with one disadvantage: such a label would only indicate the suitability of the parser in isolation and not in comparison with other parsers. Therefore, we normalize the number of correct heads for each parser (for any given UPOS-sentence) with the average of the number of correctly predicted heads across all parsers. That is, the label $y_{i,j}$ for $\text{PARSER}_{i}$ and $\text{SENT}_j$ is computed as follows:   
\begin{equation}
    y_{i,j} = \frac{\#\text{correct-heads}_{i,j}}{1/|L| \cdot \sum^{|L|}_{l=1}\# \text{correct-heads}_{l, j}}
\end{equation}
\noindent The treebanks of training languages greatly vary in size. To account for the imbalanced treebank sizes, we up-sample all below-average treebanks and down-sample all above-average treebanks. 





\subsection{ILPS Regression Model}
\label{ss:model}

 Our instance-level parser selection model is a regression model based on a Transformer architecture encoder \cite{vaswani2017attention} for UPOS-sentences. The encoding of the input UPOS-sentence is forwarded, together with the embedding vector representing the parser language, to a multi-layer percepton. It predicts the score representing the prediction of the normalized number of correct heads that the parser is expected to yield. 

\vspace{1.6mm}

\noindent \textbf{Parser and POS-tag embeddings.} 
We learn $|L|$ parser embeddings, $\{\mathbf{p}_i\}_{i = 1}^{|L|}$, one for each language  ($\text{PARSER}_i$) and $K$ embedding vectors $\{\mathbf{t}_k\}_{k = 1}^{K}$, one for each UPOS-tag \cite{petrov2012universal}. We initialize both parser and POS-tag embeddings randomly. POS-tag embeddings are then updated during the pretraining of the POS-sentence encoder. 

\vspace{1.6mm}

\noindent \textbf{UPOS-sentence encoder.}
We encode UPOS-sentences with the Transformer encoder. Let the UPOS-sentence $\text{SENT}_j = \{t^j_1, t^j_2, \dots, t^j_T\}$ be a sequence of $T$ UPOS-tags. 
We encode each token $t^i_j$ ($i \in \{1, \dots, K\}$, $j \in \{0, 1, \dots, T\}$) with a vector $\mathbf{t}^i_j$ which is the concatenation of the UPOS-tag embedding and a positional embedding for the position $j$.\footnote{We adopt the wavelength-based positional encoding from the original Transformer model \cite{vaswani2017attention}.} Let $\mathit{Transform}$ denote the encoder stack of the Transformer model \cite{vaswani2017attention} with $N_{T}$ layers, each coupling a multi-head attention net with a feed-forward net.
We then apply $\mathit{Transform}$ to the UPOS-tag sequence and obtain contextualized UPOS-tag representations as follows:   
\begin{align}
\{\mathbf{tt}^i_j\}^T_{j = 1} = \mathit{Transform}\left(\{\mathbf{t}^i_j\}^T_{j = 1}\right);
\end{align}
\noindent Following \newcite{devlin2018bert}, we pretrain the parameters of the $\mathit{Transform}$ encoder and the UPOS-tag embeddings via the masked language modeling objective on the concatenation of all training treebanks. As in the original work, we consider 15\% of randomly selected tokens in each sentence (but no more than 20 tokens) for replacement. In 80\% of the cases, we replace the UPOS-tag with the [MASK] token, in 10\% of the cases we keep the original UPOS-tag, and in remaining 10\% of the cases we replace it with a randomly chosen tag.

We fine-tune the pretrained $\mathit{Transform}$ encoder and the UPOS-tag embeddings on the main ILPS regression task. At this step, similar to \newcite{devlin2018bert}, we prepend each UPOS-sentence with a special sentence start token $t^j_0 = \text{[\textit{ss}]}$, with the aim of using the transformed representation of that token as the sentence encoding.\footnote{This eliminates the need for an additional self-attention layer for aggregating transformed token vectors into a sentence encoding. We omitted preprending the UPOS-sentences with the sentence start token in pretraining due to the lack of any sentence-level pretraining objective.} We take the transformed vector of the [\textit{ss}]  token, i.e., $\mathbf{tt}^j_0$ as the final fixed-size representation of the UPOS-sentence. 

\vspace{1.6mm}
\noindent \textbf{Feed-forward regressor and loss function.} 
For a training instance $(\text{PARSER}_i, \text{SENT}_j, y_{i,j})$, we concatenate the parser's embedding $\mathbf{p}_i$ and the UPOS-sentence encoding $\mathbf{tt}^j_0$, and feed it to a feed-forward regression network (i.e., a multi-layer perceptron, MLP), whose goal is to predict $y_{i,j}$:
\begin{equation}
    \hat{y}_{i,j} = \text{MLP}([\mathbf{p}_i;\mathbf{tt}^j_0])
\end{equation}

\noindent We define the loss function to be a simple root mean square error (RMSE) over the examples in one mini-batch as follows: 
\begin{equation}
\mathcal{L} = \sqrt{\frac{1}{N_B} \sum_{i,j}{\left(y_{i,j}-\hat{y}_{i,j}\right)^2}}
\end{equation}
\noindent where $N_B$ is the number of instances in the batch.  

\subsection{Ranking and Ensembling}
\label{sec:ranking}

We can directly use the vector of scores $\hat{\mathbf{y}_j} = \{\hat{y}_{i,j}\}_{i = 1}^{|L|}$ to rank the $|L|$ parsers according to their (predicted) parsing accuracy for the UPOS-sentence $SENT_j$ from some test treebank. 

\vspace{1.6mm}
\noindent \textbf{Pure ILPS.} 
This local parser ranking, based only on the predicted parser performance for the current UPOS-sentence $SENT_j$, is used to select one or few best parsers for that UPOS-sentence. If we select only a single best parser and only according to the instance-level predictions, we refer to the \textit{pure} instance-level parser selection (ILPS) setup:
\begin{equation}
    i_\text{ILPS}(j) \hspace{-0.2em}=\hspace{-0.2em} \underset{i}{\arg\max} \{\hat{y}_{i,j} \hspace{0.1em} | \hspace{0.1em} i \in \{1, 2, \dots, |L|\}\}
\end{equation}

\vspace{1.6mm}
\noindent \textbf{SBPS from ILPS predictions.}
ILPS predictions can be easily aggregated to produce a treebank-level estimate of the source parsers' performance for a test language. This brings the ILPS paradigm back into the single-best parser selection (SBPS) realm, hopefully with SBPS estimates originating from our ILPS predictions being more robust than competing SBPS metrics \cite{rosa-zabokrtsky-2015-klcpos3,lin-etal-2019-choosing}. For a treebank of an unseen test language consisting of $M$ POS-sentences, we get the global parser's performance estimates $\bar{y}_i$ simply by averaging ILPS predictions for that parser, $\hat{y}_{i,j}$, over all $M$ test POS-sentences:
\begin{equation}
    \bar{y}_i = \frac{1}{M} \sum_{j = 1}^{M}{\hat{y}_{i,j}}
\end{equation}
\noindent The best treebank-level parser is then selected as the one with the highest aggregate score $\bar{y}_j$:
\begin{equation}
    i_\text{SBPS\textsubscript{ILPS}} = \underset{i}{\arg\max} \{\bar{y}_i\hspace{0.1em} | \hspace{0.1em} i \in \{1, 2, \dots, |L|\}\}
\end{equation}

\vspace{1.6mm}
\noindent \textbf{Ensembling.} 
It is often the case -- both at the instance level and at the treebank level -- that two or more parsers yield similar performance. In such cases, one would expect to benefit from aggregating the predictions made by those parsers. We refer to the settings in which we consider more than one parser as ensembling (\textbf{Ens}) settings. Note that ensembling is equally applicable to both the pure ILPS setup as well as to the previously outlined SBPS\textsubscript{ILPS} setup in which we aggregate instance-level predictions to select the best ``treebank-level'' parser. In both cases, we must determine a threshold $\tau\in[0,1]$ that defines the set of ``good enough'' parsers, in relative terms w.r.t. the performance of the best parser. The sets of parsers whose trees are to be merged are obtained as follows: 
\begin{align}
  \{i_\text{ILPS}\}_\tau(j) &= \{i \hspace{0.05em}| \hspace{0.05em} \forall i : \hat{y}_{i,j} \geq \max(\hat{y}_{i,j}) \cdot \tau \} \label{eq:sel_ilps}, \\
  \{i_\text{SBPS\textsubscript{ILPS}}\}_\tau &= \{i \hspace{0.1em} | \hspace{0.1em} \forall i : \bar{y}_{i} \geq \max(\bar{y}_{i}) \cdot \tau \} \label{eq:sel_sbps}. 
\end{align}
\noindent where Eq~\eqref{eq:sel_ilps} refers to the pure ILPS setting, and Eq.~\eqref{eq:sel_sbps} refers to the SBPS\textsubscript{ILPS} setting.

\subsection{Reparsing}
\label{sec:reparsing}

After selecting multiple parsers in the ensemble settings, we need to merge their produced parse trees into a final tree. It is commonly referred to as \textit{reparsing} \cite{sagae-lavie-2006-parser}. We resort to a standard reparsing procedure in which we: (1) merge the trees produced by individual parsers into a weighted graph $G$ -- the parser $i$ contributes to an edge with the weight $w_i = \hat{y}_{i,j}$ (for pure ILPS; for SBPS\textsubscript{ILPS}, $w_i = \bar{y}_{i}$) if the parser $i$ predicted that edge, and with $w_i = 0$ otherwise; (2) induce the Maximum Spanning Tree (MST) of $G$ \cite{edmonds1967optimum} as the final parse of the input UPOS-sentence, see again Figure~\ref{fig:overview}.

\section{Experimental Setup}
\noindent \textbf{Data.} 
We perform all experiments on the UD v2.3 dataset,\footnote{\url{https://universaldependencies.org/}} as it contains a wide array of both resource-rich languages with large treebanks -- split into train, development, and test portions -- and low-resource languages with small test treebanks. For our experiments, we select 42 languages with the largest treebanks as our resource-rich source languages for training, and a set of 20 typologically diverse low-resource languages for testing.\footnote{We provide the full list of languages with the corresponding treebank sizes in the supplementary material.} Following established practice \cite{wang-eisner-2018-synthetic}, at inference we use gold UPOS-tags of test treebanks for all models in comparison.\footnote{While this does not affect the fairness of model comparisons (since all models, including baselines, are exposed to gold UPOS-tags), it does render reported results as models' upper bounds w.r.t. the realistic low-resource setting in which one would resort to noisier, automatically induced UPOS-tags.} 



\vspace{1.6mm}
\noindent \textbf{ILPS Hyperparameters} are optimized via fixed-split cross-validation on our training set (see \S\ref{sec:prep_tr_data}). We set the embedding size for both parser embeddings and UPOS-tag embeddings, as well as the hidden size of the feed-forward Transformer layers to $256$. The transformer encoder has $N_T = 3$ layers with $8$ attention heads in each layer. 
We update the model in mini-batches of $16$ examples, using Adam \cite{kingma2014adam} with the default parameters: $\beta_1=0.9$, $\beta_2=0.999$, and $\epsilon=10^{-8}$, with an initial learning rate set to $10^{-4}$. The regression MLP has $2$ hidden layers with $256$ units each, plus a linear projection layer that compresses the $256$-dimensional vector into a single prediction score. We perform early stopping based on the loss on the development set. For all ensembles, we set the parser inclusion threshold $\tau$ to $0.9$. 

\vspace{1.6mm}

\noindent \textbf{Oracle scores and baselines.} In order to provide more context for the reported ILPS scores, we also report the results of two \textit{oracle} methods described in \S\ref{sec:motivation}: the oracle single-best parser selection (\textsc{Or-SBPS}), and the oracle instance-level best parser selection (\textsc{Or-ILPS}). 
We compare to three competitive baselines: (1) the standard multi-source parser (\textsc{MSP}) baseline which trains a single parser model on the concatenation of all training treebanks;\footnote{We have run two variants of the multi-source model (MSP): a) \textit{balanced} (trained on the treebanks downsampled or upsampled to the average treebank size as done in \S\ref{sec:prep_tr_data}); b) \textit{all} (trained on the concatenation of the full treebanks without any adjustment). For brevity, we report the results only with the latter, as it produced stronger overall performance.} and two competitive SBPS baselines, (2) \textsc{KL-SBPS} -- treebank-level parser selection based on the Kullback-Leibler divergence between UPOS-tag trigram distributions of the source and target language treebanks \cite{rosa-zabokrtsky-2015-klcpos3} and (3) \textsc{L2V-SBPS} -- treebank-level parser selection based on the cosine similarity between the syntax-based vectors of the source and target language from WALS \cite{lin-etal-2019-choosing}.  

\vspace{1.6mm}

\noindent \textbf{Ensembles.} We evaluate two ensembles based on the predictions of our ILPS-based regression model, described in \S\ref{sec:ranking}: (1) an instance-level ensemble in which we merge the trees of the best parsers for each sentence (\textsc{Ens-ILPS}) and (2) \textsc{Ens-SBPS\textsubscript{ILPS}} -- an ensemble merging the trees of treebank-level best parsers, where the treebank-level estimates are aggregated from the instance-level predictions). We evaluate comparable ensembles (i.e., with the same parser inclusion performance threshold $\tau = 0.9$) for both SBPS baselines: \textsc{Ens-KL-SBPS} and \textsc{Ens-L2V-SBPS}.

\section{Results and Discussion}

\begin{figure}
    \centering
    \includegraphics[width=\linewidth]{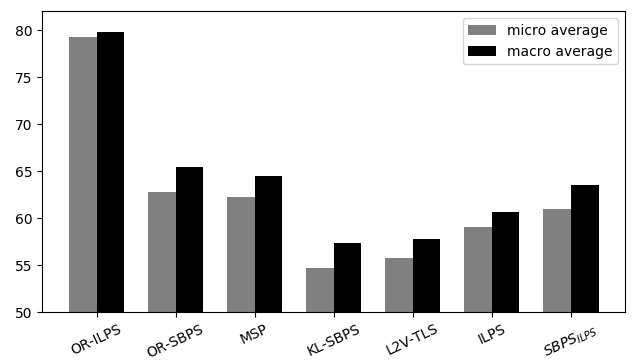}
    \caption{Performance (UAS) for single-parser selection models, micro- and macro- averaged, respectively, across 20 test languages.}
    \label{fig:aggregateresults_singleparser}
    \vspace{-1mm}
\end{figure}
\begin{figure}
    \centering
    \includegraphics[width=\linewidth]{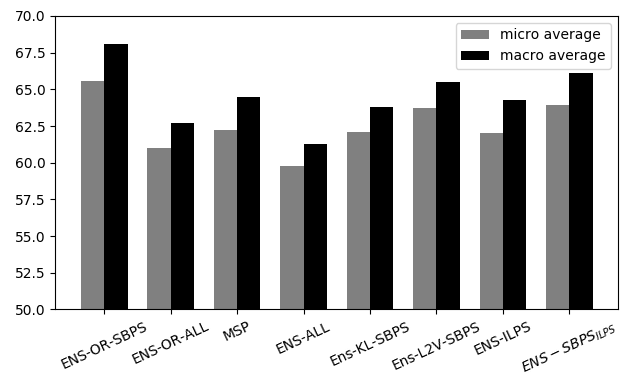}
    \caption{Performance (UAS) for ensemble (i.e., few-parser selection) models, micro- and macro- averaged, respectively, across 20 test languages.}
    \label{fig:aggregateresults_ensemble}
    \vspace{-1mm}
\end{figure}

We first show the results for single-best parser selection models. We then proceed to a more realistic \textit{ensemble} setup in which the models are allowed to select more than just one parser. 


\vspace{1.6mm}
\noindent \textbf{Single-parser selection.} 
We report results (UAS) for all single-parser selection methods (i.e., no ensembles) along with the oracle scores on all 20 test treebanks. Table~\ref{tbl:42lsetup-nonensemble} provides performance per language, and Figure~\ref{fig:aggregateresults_singleparser} shows the summary of the results.
\setlength{\tabcolsep}{1.5pt}
\begin{table*}[!t]
\centering
\def\arraystretch{0.97}
{\scriptsize
{
\begin{tabularx}{\linewidth}{l X X X X X X X X X X X X X X X X X X X X X X}
\toprule
& am & be & bm & br & bxr & cop & fo & ga & hsb & hy & kk & lt & mt & myv & sme & ta & te & th & yo & yue & \textbf{Ma} & \textbf{Mi}  \\ \midrule
\multicolumn{20}{l}{\textit{Oracles}} \\ \midrule
\textsc{Or-ILPS} & 79.7 & 87.3 & 82.8 & 83.1 & 74.2 & 86.3 & 71.9 & 76.8 & 86.1 & 75.2 & 81.2 & 74.7 & 85.9 & 83.5 & 79.3 & 73.8 & 94.8 & 70.3 & 71.7 & 78.0 & 79.8 & 79.3\\
\textsc{Or-SBPS} & 62.1 & 77.4 & 66.9 & 61.3 & 54.5 & 76.3 & 52.1 & 65.2 & 74.8 & 62.8 & 64.4 & 65.9 & 74.2 & 67.1 & 58.5 & 63.6 & 83.2 & 57.6 & 60.2 & 58.2 & 65.4 & 62.8 \\ \midrule
\multicolumn{20}{l}{\textit{Baselines}} \\ \midrule
\textsc{MSP} & 56.7 & \textbf{78.7} & \textbf{70.7} & \textbf{62.8} & \textbf{57.8} & \textbf{77.2} & \textbf{51.4} & \textbf{66.3} & \textbf{78.2} & \textbf{67.9} & \textbf{64.9} & \textbf{63.3} & \textbf{77.4} & 65.3 & \textbf{61.4} & 59.5 & 75.5 & 56.3 & \textbf{59.0} & 37.6 & \textbf{64.5} & \textbf{62.2} \\
\textsc{KL-SBPS} & 47.5 & 77.1 & 54.2 & 56.2 & 48.9 & 65.9 & 48.7 & 65.2 & 72.9 & 62.8 & 57.5 & 46.8 & 69.0 & 54.3 & 57.2 & 47.3 & 74.8 & 35.2 & 47.9 & \textbf{56.7} & 57.3 & 54.7\\
\textsc{L2V-TLS} & 26.9 & 70.4 & 55.6 & 58.7 & 53.8 & 69.9 & 48.9 & 61.3 & 71.1 & 57.5 & 64.4 & 49.9 & 70.4 & \textbf{67.1} & 57.2 & 54.5 & \textbf{83.2} & 47.6 & 45.2 & 41.7 & 57.8 & 55.7  \\
\midrule
\multicolumn{20}{l}{\textit{ILPS models (\textit{ours})}} & \\ \midrule
ILPS & 57.1 & 75.3 & 60.2 & 60.5 & 53.5 & 70.9 & 49.5 & 62.2 & 72.5 & 56.4 & 58.6 & 56.2 & 71.7 & 61.2 & 55.1 & 58.0 & 71.4 & 52.9 & 54.9 & 53.5 & 60.6 & 59.0 \\
SBPS\textsubscript{ILPS} & \textbf{62.1} & 77.4 & 62.7 & 60.5 & 54.5 & 75.3 & 48.6 & 61.4 & 72.9 & 62.8 & 64.1 & 58.6 & 73.2 & 67.1 & 57.2 & \textbf{63.6} & 77.0 & \textbf{57.6} & 57.2 & 56.2 & 63.5 & 61.0 \\ 
\bottomrule
\end{tabularx}
}}
\vspace{-1mm}
\caption{Results for single-parser selection models. Results for 42 parsers (an exception is the MSP model which trains a single parser on the concatenation of all training treebanks) on 20 low-resource test languages. Ma \& Mi: average performance across 20 languages, macro- and micro-averaged scores, respectively. The best result in each column, not considering oracle scores, is in bold.}
\label{tbl:42lsetup-nonensemble}
\vspace{0mm}
\end{table*}
%
\noindent Our \textit{pure} instance-based parser selection model (ILPS) significantly\footnote{Significance tested with the Student's two-tailed t-test at $p = 0.01$ for sets of sentence-level UAS scores.} outperforms both SBPS baselines (KL-SBPS and L2V-SBPS) averaged across all languages (see Figure~\ref{fig:aggregateresults_singleparser}). Individual instance-level predictions made by \textsc{ILPS}, however, do seem to be rather noisy. This is supported by the observation that SBPS\textsubscript{ILPS} significantly outperforms ILPS. Since SBPS\textsubscript{ILPS} is a simple treebank-level aggregation of ILPS sentence-level predictions, the gain can only be explained as the product of noise elimination through aggregation. \textsc{ILPS} outperforms KL-SBPS and L2V-SBPS on 13/20 and 14/20 test languages, respectively, whereas SBPS\textsubscript{ILPS} improves on 17/20 and 16/20 languages over the respective baselines. This first set of results, the preliminary comparison with well-established and competitive baselines for delexicalized parser transfer, seems encouraging and validates the viability of the instance-based parser selection paradigm. 

\textsc{ILPS} and SBPS\textsubscript{ILPS} still do not match the performance of the multi-source parser (MSP) in this simple single-parser-selection setup. We find this somewhat expected: ILPS and SBPS\textsubscript{ILPS} are based on parsers trained on single treebanks, whereas MSP is trained on the concatenation of all training treebanks. Therefore, we include MSP as a baseline in our ensemble evaluation as well.  
%
%


\vspace{1.6mm}
\noindent \textbf{Ensemble evaluation results.} 
We show the results for the ensemble models in Table~\ref{tbl:42lsetup_ensemble}. A summary of results for this setup is provided in Figure~\ref{fig:aggregateresults_ensemble}.
\setlength{\tabcolsep}{1.5pt}
\begin{table*}[!t]
\centering
\def\arraystretch{0.97}
{\scriptsize
{
\begin{tabularx}{\linewidth}{l X X X X X X X X X X X X X X X X X X X X X X}
\toprule
& am & be & bm & br & bxr & cop & fo & ga & hsb & hy & kk & lt & mt & myv & sme & ta & te & th & yo & yue & \textbf{Ma} & \textbf{Mi} \\ \midrule
\multicolumn{20}{l}{\textit{Oracle ensembles}} \\ \midrule
\textsc{Ens-Or-SBPS} & 62.9 & 79.2 & 70.3 & 64.2 & 62.1 & 78.9 & 50.5 & 68.6 & 78.5 & 66.3 & 69.2 & 65.9 & 78.3 & 66.8 & 64.3 & 65.3 & 83.9 & 61.8 & 62.8 & 61.7 & 68.1 & 65.6\\
\textsc{Ens-Or-All} & 59.6 & 78.0 & 70.8 & 63.5 & 49.8 & 78.4 & 50.6 & 67.6 & 76.2 & 58.6 & 52.6 & 56.2 & 76.6 & 64.5 & 52.3 & 48.0 & 67.4 & 59.8 & 62.2 & 61.3 & 62.7 & 61.0 \\ \midrule
\multicolumn{20}{l}{\textit{Baseline ensembles}} \\ \midrule
\textsc{MSP} & 56.7 & 78.7 & 70.7 & 62.8 & 57.8 & 77.2 & \textbf{51.4} & 66.3 & \textbf{78.2} & \textbf{67.9} & 64.9 & \textbf{63.3} & \textbf{77.4} & 65.3 & \textbf{61.4} & 59.5 & 75.5 & 56.3 & 59.0 & 37.6 & 64.5 & 62.2 \\
\textsc{Ens-All} & 59.2 & 77.1 & 70.5 & 63.0 & 45.6 & 78.2 & 50.3 & 67.2 & 75.4 & 57.4 & 47.5 & 56.0 & 76.2 & 64.1 & 52.0 & 38.4 & 66.2 & 58.3 & 61.6 & \textbf{61.6} & 61.3 & 59.8 \\
Ens-KL-SBPS & \textbf{60.7} & \textbf{79.2} & \textbf{71.2} & 63.5 & 56.6 & 78.1 & 50.6 & 67.7 & 76.0 & 57.2 & 58.6 & 53.5 & 76.5 & 65.3 & 53.2 & 58.8 & 68.4 & 57.4 & \textbf{62.2} & 61.1 & 63.8 & 62.1 \\
Ens-L2V-SBPS & 60.4 & 78.7 & 68.9 & \textbf{63.8} & \textbf{61.5} & 77.5 & 50.7 & 68.1 & 76.7 & 59.1 & \textbf{70.7} & 54.5 & 77.0 & 65.2 & 50.9 & \textbf{66.5} & 74.3 & 60.2 & 61.6 & 62.8 & 65.5 & 63.7\\ \midrule
\multicolumn{20}{l}{\textit{ILPS model-based ensembles (\textit{ours})}} & \\ \midrule
\textsc{Ens-ILPS} & 59.6 & 78.7 & 68.2 & 62.8 & 56.1 & 77.9 & 50.8 & 67.2 & 76.5 & 60.8 & 61.7 & 60.6 & 76.5 & 63.4 & 56.5 & 61 & 72.8 & 57.4 & 60.0 & 57.4 & 64.3 & 62.0 \\
\textsc{Ens-SBPS\textsubscript{ILPS}} & 60.0 & 78.7 & 70.8 & \textbf{63.8} & 61.0 & \textbf{78.4} & 50.5 & \textbf{68.2} & 77.5 & 58.9 & 68.1 & 62.9 & 76.7 & \textbf{66.8} & 53.6 & 65.3 & \textbf{78.5} & \textbf{60.5} & 62.0 & 60.1 & \textbf{66.1} & \textbf{63.9} \\
\bottomrule
\end{tabularx}
}}
\vspace{-1mm}
\caption{Results for ensemble-based parser selection models. Additional models: \textsc{Ens-Or-All} -- merges parses by all 42 parsers, but uses oracle performance as parser weights; \textsc{Ens-All} -- ensembles all 42 parsers, with equal weights. An exception is the \textsc{MSP} model which is not an ensemble model, but rather trains a single parser on the concatenation of all training treebanks. Ma \& Mi: average
performance across 20 languages, macro- and micro-averaged scores, respectively. The best result in each column, not considering oracle scores, is in bold.}
\label{tbl:42lsetup_ensemble}
\vspace{-1.5mm}
\end{table*}
Allowing for the selection of more than a single parser in cases in which our ILPS-based predictions warrant so (i.e., when two or more parsers yield similarly good performance for some low-resource language) allows SBPS\textsubscript{ILPS} (i.e., its ensemble version, Ens-SBPS\textsubscript{ILPS}) to significantly outperform the strong MSP baseline. The two SBPS baseline methods in their ensemble variants (Ens-KL-SBPS and Ens-L2V-SBPS) reduce the gap in comparison with the previous single-parser selection setup (see Table~\ref{tbl:42lsetup-nonensemble} again). However, our treebank-level parser selection model based on instance-level predictions (Ens-SBPS\textsubscript{ILPS}) still significantly outpeforms the ensembles of the other two SBPS methods.

As en encouraging finding, both Ens-ILPS and Ens-SBPS\textsubscript{ILPS} outperform the \textit{oracle} baseline Ens-Or-All, which merges parses produced by all training parsers, using their gold performance on the test treebanks for weighting the individual parser contributions. Furthermore, Ens-SBPS\textsubscript{ILPS} also improves over the oracle single-parser selection \textsc{or-SBPS} reported in Table~\ref{tbl:42lsetup-nonensemble}. In summary, we believe these results provide sufficient evidence for the viability of the ILPS transfer paradigm and warrant further research efforts in this direction.  

\section{Related Work}

Parsing languages with no training data has been a very active topic of research for nearly a decade since the pivotal works by~\citet{mcdonald-etal-2011-multi} and~\citet{petrov2012universal}. Many diverse approaches are explored along the lines of model transfer, annotation projection, machine translation \cite{Tackstrom:2013naacl,Guo:2015acl,Zhang:2015emnlp,tiedemann2016synthetic,rasooli2017cross}, and selective sharing based on language typology \cite{Naseem:2012acl} and structural similarity \cite{Ponti:2018acl}. However, vast majority of prior work involves bulk evaluation, whereby transfer parsers are validated by mean accuracy on test data. Such evaluation protocols stand in contrast with the fact that languages exhibit high variance in syntactic structure, which calls for a sensitive treatment of \textit{every sentence}. While an oracle single-source parser may be appropriate for the majority of sentences in a given dataset, instance-based treatment closes the gap to the best achievable result given an array of pretrained parsers, as we also show in \S\ref{sec:motivation}.

Early efforts in this line of research include data point selection where language models are used to capture the prevalent syntactic structure of a language and score potential training instances such that a multi-source parser is trained on the mixture of training instances that are most similar to the test language instances \cite{sogaard2011data}.  Instead of instance selection one can also apply instance reweighting in accordance to their similarity to the test language~\cite{sogaard-wulff-2012-empirical}. Regardless of whether we attempt to align languages on an instance-level or on a treebank-level there is a need for a similarity measure between languages. Prior work relied on existing manually curated resources such as the URIEL database~\cite{littell2017uriel}, using the KL-Divergence on POS-trigrams~\cite{rosa-zabokrtsky-2015-klcpos3}, or handcrafted features derived from the datasets at hand.

Our work is most similar to the recent work of ~\citet{lin-etal-2019-choosing}: they learn to score and rank languages in order to predict the top transfer languages. However, contrary to their work, our approach does not employ a model to learn the ranking, but transforms the labels to directly reflect the ranking when we train the scoring model. In addition, we stress the importance of instance-based learning for cross-lingual parser transfer in particular. Another core difference is that our approach is an end-to-end system without external resources or handcrafted static features. Instead, our framework relies on trainable parser embeddings that encode the necessary features in a single representation.

From another viewpoint, the work of~\citet{wang-eisner-2016-galactic,wang-eisner-2018-surface,wang-eisner-2018-synthetic} explores the potential of synthesizing and reordering delexicalized POS sequences to come up with better parser transfer without unrealistic assumptions on target-language resources. Their work in synthetic delexicalization is compatible with ours as it lends itself entirely to instance-based parsing. Finally, the line of work by~\citet{ammar-etal-2016-many} in learning monolithic models over multiple training languages, and its continuation for zero-shot learning by~\citet{kondratyuk-straka-2019-75} also promises to abstract away from language boundaries, but still records significantly lower zero-shot scores than our proposal.

\section{Conclusion and Future Work}

In this work, we indicated that there is a large disparity between mean test-set and per-instance accuracy in cross-lingual parser transfer setups. We showed convincing evidence that one source parser is not the optimal choice for all target-language sentences. Motivated by the analysis, we proposed a novel approach to close this gap: instance-based parser selection. Our framework provides competitive results, where in the ensemble setting we outperform all baselines, and markedly even the single-source oracle parser selection, while using a simple thresholding heuristic to select the parsers.

We see the proposed model as the first exploratory step in the direction of robust instance-level parser transfer, which opens several avenues for future research. While this proof-of-concept work assumed the existence of gold POS tags, we will also experiment with the same approach ``in the wild'', with learned or transferred POS taggers, and we will also extend the study to lexicalized parser transfer. Future work may also include learning to rank parsers instead of applying simple heuristics. Further improvements may be obtained by using the accuracy predictions from our model as a feature and combining it with external linguistic features~\cite{Ponti:2019cl}. The idea of instance-based parser selection lends itself also to domain adaptation settings, following~\citet{plank2011effective}, even for well-resourced languages.


\section*{Acknowledgments}
IV is supported by the ERC Consolidator Grant LEXICAL: Lexical Acquisition Across Languages (no 648909).

\bibliography{references}
\bibliographystyle{acl_natbib}

\clearpage
{\small
\begin{table}[h]
    \centering
    \begin{tabular}{l|c|c}
    \toprule 
Train Language & \#sentences & \#tokens \\
\hline
Estonian (et) & 24384 & 341122 \\
Korean (ko) & 23010 & 296446 \\
Latin (la) & 16809 & 293306 \\
Norwegian (no) & 15696 & 243887 \\
Finnish (fi) & 14981 & 127602 \\
French (fr) & 14450 & 354699 \\
Spanish (es) & 14305 & 444617 \\
German (de) & 13814 & 263804 \\
Polish (pl) & 13774 & 104750 \\
Hindi (hi) & 13304 & 281057 \\
Catalan (ca) & 13123 & 417587 \\
Italian (it) & 13121 & 276019 \\
English (en) & 12543 & 204585 \\
Dutch (nl) & 12269 & 186046 \\
Czech (cs) & 10160 & 133637 \\
Portuguese (pt) & 9664 & 255755 \\
Bulgarian (bg) & 8907 & 124336 \\
Slovak (sk) & 8483 & 80575 \\
Romanian (ro) & 8043 & 185113 \\
Latvian (lv) & 7163 & 113405 \\
Japanese (ja) & 7133 & 160419 \\
Croatian (hr) & 6983 & 154055 \\
Slovenian (sl) & 6478 & 112530 \\
Arabic (ar) & 6075 & 223881 \\
Basque (eu) & 5396 & 72974 \\
Ukrainian (uk) & 5290 & 88043 \\
Hebrew (he) & 5241 & 137721 \\
Persian (fa) & 4798 & 121064 \\
Indonesian (id) & 4477 & 97531 \\
Danish (da) & 4383 & 80378 \\
Swedish (sv) & 4303 & 66645 \\
Urdu (ur) & 4043 & 108690 \\
Chinese (zh) & 3997 & 98608 \\
Russian (ru) & 3850 & 75964 \\
Turkish (tr) & 3685 & 37918 \\
Serbian (sr) & 2935 & 65764 \\
Galician (gl) & 2272 & 79327 \\
Greek (el) & 1662 & 42326 \\
Uyghur (ug) & 1656 & 19262 \\
Vietnamese (vi) & 1400 & 20285 \\
Afrikaans (af) & 1315 & 33894 \\
Hungarian (hu) & 910 & 20166 \\
\hline
Average (avg) & 8483 & 158233 \\
\bottomrule
    \end{tabular}
    \caption{42 source languages on which we trained monolingual parsers. Number of sentences (\#sentences) and total token count (\#tokens).} .
    \label{tab:trainStats}
\end{table}}

\begin{table}[h]
    \centering
    \begin{tabular}{l|c|c}
    \toprule
    Test Language & \#sentences & \#tokens \\
    \hline
Erzya (myv) & 1550 & 15790 \\
Faroese (fo) & 1208 & 10002 \\
Amharic (am) & 1074 & 10010 \\
Kazakh (kk) & 1047 & 10007 \\
Bambara (bm) & 1026 & 13823 \\
Thai (th) & 1000 & 22322 \\
Buryat (bxr) & 908 & 10032 \\
Breton (br) & 888 & 10054 \\
North Sami (sme) & 865 & 10010 \\
Cantonese (yue) & 650 & 6264 \\
Upper Sorbian (hsb) & 623 & 10736 \\
Maltese (mt) & 518 & 11073 \\
Armenian (hy) & 470 & 11438 \\
Irish (ga) & 454 & 10138 \\
Coptic (cop) & 267 & 6541 \\
Telugu (te) & 146 & 721 \\
Tamil (ta) & 120 & 1989 \\
Yoruba (yo) & 100 & 2666 \\
Belarusian (be) & 68 & 1382 \\
Lithuanian (lt) & 55 & 1060 \\
\hline 
Average & 1094 & 8802 \\
\bottomrule
    \end{tabular}
    \caption{20 unseen test languages. Number of sentences (\#sentences) and total token count (\#tokens).}
    \label{tab:testStats}
\end{table}


\end{document}